\documentclass[letterpaper, 10pt, conference]{ieeeconf}  

\IEEEoverridecommandlockouts                              

\usepackage{amsmath} 
\usepackage{amssymb}  
\usepackage{graphicx}                  
\usepackage{url}                       
\usepackage[subrefformat=parens]{subcaption}
\usepackage{siunitx}
\usepackage{algorithm}
\usepackage{algpseudocode}
\usepackage{siunitx}

\graphicspath{
  {./fig/}
}

\RequirePackage{xspace}

\newcommand{\etal}{\emph{et al.}\xspace}

\newcommand{\apriori}{\emph{a priori}\xspace}
\newcommand{\stimes}{{\times}}

\newcommand\algoline[1]{[line\,\ref{#1}]\xspace}
\newcommand\algolines[2]{[lines\,\ref{#1}--\ref{#2}]\xspace}
\newcommand{\selfnav}{\textit{self-nav}\xspace}
\newcommand{\fieldname}[1]{\textit{#1}\xspace}
\newcommand{\insone}{\fieldname{simple-loop}}
\newcommand{\instwo}{\fieldname{two-bridge}}
\newcommand{\insthree}{\fieldname{two-loop}}

\title{\LARGE \bf
  Active Modular Environment for Robot Navigation
}

\author{Shota Kameyama$^1$, Keisuke Okumura,$^1$ Yasumasa Tamura$^1$ and Xavier D\'{e}fago$^1$
 \thanks{
   $^{1}$The authors are with School of Computing, Tokyo Institute of Technology, Tokyo, Japan.
   {\scriptsize\tt\{ kameyama.s, okumura.k, tamura.y, defago.x \} @coord.c.titech.ac.jp}.
 }
}

\begin{document}

\maketitle
\thispagestyle{empty}
\pagestyle{empty}

\begin{abstract}
  This paper presents a novel robot-environment interaction in navigation tasks such that robots have neither a representation of their working space nor planning function, instead, an active environment takes charge of these aspects.
  This is realized by spatially deploying computing units, called cells, and making cells manage traffic in their respective physical region.
  Different from stigmegic approaches, cells interact with each other to manage environmental information and to construct instructions on how robots move.

  As a proof-of-concept, we present an architecture called \emph{AFADA} and its prototype, consisting of modular cells and robots moving on the cells.
  The instructions from cells are based on a distributed routing algorithm and a reservation protocol.
  We demonstrate that AFADA achieves efficient robot moves for single-robot navigation in a dynamic environment changing its topology with a stochastic model, comparing to self-navigation by a robot itself.
  This is followed by several demos, including multi-robot navigation, highlighting the power of offloading both representation and planning from robots to the environment.
  We expect that the concept of AFADA contributes to developing the infrastructure for multiple robots because it can engage online and lifelong planning and execution.
\end{abstract}
\section{Introduction}\label{lab:sec:introduction}
Representation, planning, execution, and their smooth integration are essential factors for developing intelligent systems.
In particular, representation, i.e., how to abstract and model the world, has been a central issue for AI and robotics~\cite{davis1993knowledge};
however, there is room to consider whether robots themselves should have a representation of their working environment.
As famously argued by Brooks~\cite{brooks1991intelligence}, another view is to ``use the world as its own model''.

\paragraph*{Navigation}
We can see derivative concepts, direct use of the world as the representation, in navigation tasks.
Navigation is a fundamental robotics challenge that enables autonomous robots to reach their destinations.
In general, robots use internal maps as a representation of the environment;
however, this entails some serious difficulties, still making navigation challenging~\cite{khaliq2015stigmergy}, e.g.,
accurate and robust self-localization,
working in a dynamic environment where maps are updated frequently,
the necessity of a map whenever robots enter a new workspace, and lack of consistency between the internal maps of different robots in multi-robot scenarios.
Such problems mostly stem from discrepancies between external physical objects and internal representation.

\begin{figure}
  \centering
  \includegraphics[width=1.0\linewidth]{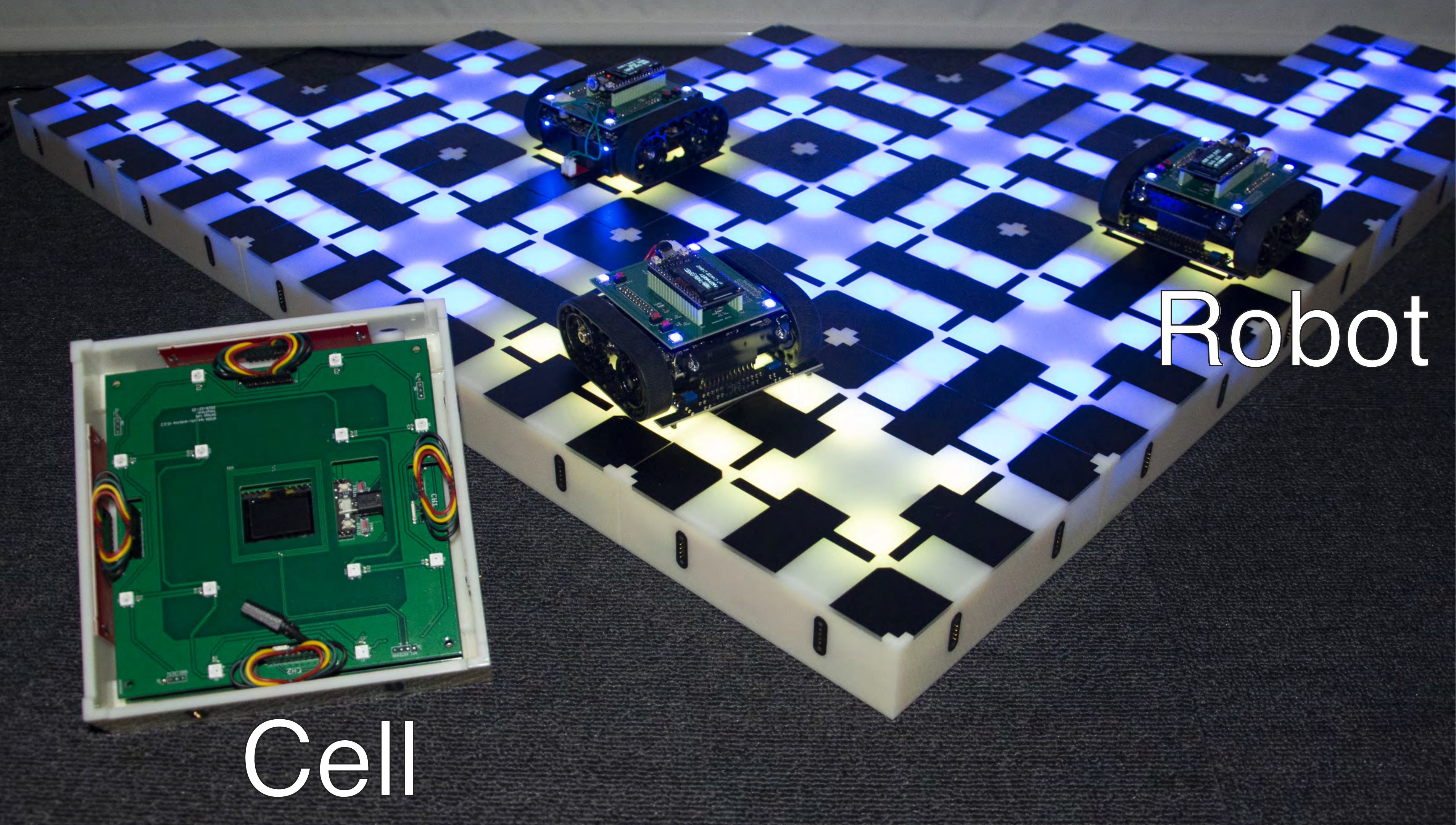}
  \caption{\textbf{AFADA prototype.}}
  \label{fig:overview}
\end{figure}

\paragraph*{Passive environment}
The use of the environment itself as the representation can be a powerful tool for navigation.
This is realized by spatially deploying sensors or tags.
In such workspaces, robots without explicit internal maps can navigate individually.
%
For instance, in the early development of self-driving cars, they were guided by electronic devices embedded in roadways~\cite{bimbraw2015autonomous}.
%
Sensor networks can help navigation, e.g., Verma \etal~\cite{verma2005selection} used a static sensor network to guide a robot in an unknown environment.
Signals from RFID tags surrounding the environment have been used to navigate a robot~\cite{gueaieb2008intelligent,kim2008direction,park2009autonomous}.
%
In nature, stigmergy~\cite{theraulaz1999brief}, a mechanism of indirect coordination through the environment between agents, achieves unexpected collective behaviors such as coordination in social insects~\cite{camazine2003self}.
Stigmergic approaches in multi-robot systems have been popular, e.g.,~navigating robots by artificial pheromones dropped into the environment, taking the form of chemical compounds, embedded RFID tags, etc~\cite{johansson2009navigating,fujisawa2014designing,khaliq2015stigmergy}.
%
These studies can also be seen as examples of spatial computing~\cite{zambonelli2004spatial}.

\paragraph*{Active environment}
This paper aims at offloading not only the representation that robots have of the environment, but also the planning function, aiming at a smooth integration of representation, planning, and execution.
In navigation tasks, the offloading is embodied as follows.
A robot has a destination but does not know the environment or its own position.
Instead, the environment actively and repeatedly issues instructions, i.e., planning, instructing the robot where to move.
This is realized by deploying static computing units (neither simple sensors nor tags) and by coordinating between these units.

\paragraph*{Rationale}
The rationale of this concept addresses the following three aspects:
1)~\emph{functional separation}:
Robots are relieved from the burden of collecting relevant information and planning their trajectories, letting them focus on other tasks.
This also implies that even robots lacking expensive sensors can navigate the environment.
2)~\emph{response to dynamic environment}:
The system allows for faster response to unexpected environmental changes because the environment itself does planning.
E.g.,~if a road is blocked by some accident, detours can be computed at an early timing, saving wasted time for robots.
3)~\emph{multi-robot coordination}:
When several robots are working in the same space, robots have to pay attention to each others to prevent collisions.
In our system, the environment manages the robots' locations in real-time and can plan collision-free trajectories.

Without the direct use of the environment as the representation, it is common to separate agents between planning and execution, especially in multi-agents/robots scenarios.
%
Centralized approaches are often used in cooperative multi-agent planning to give plans to distributed agents~\cite{torreno2017cooperative}.
%
In the field of intersection management for autonomous vehicles, one major approach is that a coordination unit manages trajectories in the intersection~\cite{dresner2008multiagent,chen2015cooperative}.
%
In an automated warehouse with hundreds of robots conveying packages~\cite{wurman2008coordinating}, planning problems where a centralized unit plan paths for all agents have been actively studied~\cite{ma2017overview,salzman2020research}.

\paragraph*{Contributions}
As a proof-of-concept of offloading both representation and planning function from robots to the environment, we present \emph{AFADA}\footnote{
  AFADA: Adaptive Fully Automated Decentralized Architecture.
} and its prototype;
an architecture that consists of mobile robots that evolve over an active environment made of flat cells each equipped with a computing unit (see Fig.~\ref{fig:overview}).
The prototype consists of modular cells designed from scratch and robots built around a Zumo robot base.
Each cell can communicate with adjacent cells and a robot on the cell.
Using local communication, cells collectively manage the environmental information such as locations of robots or routing information for navigation, despite the addition, removal, or even after the failure of cells.
In multi-robot scenarios, collision avoidance is achieved by coordination between cells.
Robots just follow instructions from the cells, i.e.,~robots necessitate neither representation nor planning.
We empirically demonstrate that AFADA achieves efficient robot moves through single-robot navigation in a dynamic environment changing its topology according to a stochastic model.
Several demos, including multi-robot navigation, highlight the benefits of offloading.

\paragraph*{Other related work}
%
Closer to our aspirations, the Kilogrid~\cite{valentini2018kilogrid} is a modular environment consisting of computing nodes arranged in a grid with centralized control, making it easier to experiment to collect data with large groups of robots, Kilobot~\cite{rubenstein2014programmable}.
%
Johnson and Mitra~\cite{johnson2015safe} studied a theoretical model of distributed traffic control where a fixed environment consists of cells that guide mobile entities from predetermined sources to targets.

\paragraph*{Paper organization}
The paper is structured as follows.
Section~\ref{sec:system} states system assumptions and formulates the navigation problem.
Section~\ref{sec:elements} presents main logical  aspects of AFADA:  construction of routing tables, as well as integrated planning and execution.
Section~\ref{sec:prototyping} presents the hardware architecture.
Section~\ref{sec:evaluation} evaluates single-robot navigation in a dynamic environment.
This is followed by several demos in Section~\ref{sec:case-studies}.
Finally, Section~\ref{sec:discussion} concludes the paper with a discussion of future directions.

\section{System Assumptions and Problem}
\label{sec:system}
The system consists of two kinds of actors: \emph{cells} and \emph{robots}. The system assumes no prior knowledge of robots or cells, even in a dynamic environment or in multi-robots scenarios. This section explains the assumptions and formulates the problem of navigation in AFADA.

\paragraph*{Cells} are computing units deployed spatially which collectively form the environment in two ways: a)~as a two-dimensional physical grid and b)~ as a communication network.
All cells are shaped identically and each cell manages its own square area, such that areas never overlap but borders can be shared (i.e., when two cells are in physical contact).
Cells can communicate with adjacent cells, thus forming a network.
Cells constitute a graph $G$, representing both the physical grid and the communication network.
We assume that cells have a \emph{unique id} but are otherwise identical.

\paragraph*{Robots} evolve over $G$.
Their size is smaller than one cell.
A robot can move on cells along edges, i.e, the robot occupies at most two cells simultaneously.
Two robots occupying the same cell are regarded as colliding; a situation which must be avoided.
When a robot is on a single cell, it can communicate with the cell wirelessly and request or receive instructions.

Neither cells nor robots know $G$ and or the set of robots $R$ \apriori.
Furthermore, we assume that $G$ changes dynamically, in response to the addition, removal, or crash of cells.
The set of robots $R$ is also dynamic, in the sense that, robots can enter or leave the system.

\paragraph*{Navigation}
Each robot is assigned a list of destination cells.
The navigation problem consists in making robots visit all destinations in their list. The problem has many variants, such as, whether it requires to visit the destinations in the given sequence, or whether all goal must initially exist in $G$.

Note that, because it is common to use a discrete representation of the environment in conventional navigation problems, the problem defined here is adaptive to a wide range of problems by regarding a cell as just a vertex.

\section{Elements of AFADA}
\label{sec:elements}
This section presents two key logical aspects of AFADA: \emph{representation} and its maintenance, and \emph{integration} of planning and execution.

\subsection{Representation}
In AFADA, the representation of the environment is stored and maintained collectively in cells, in the form of \emph{routing tables} maintained at each cell.
Similar to \emph{packet routing} in communication networks, each intermediate cell guides the robot to the neighbor cell that will bring it closer to its final destination. Routing tables maintain the information necessary to make these decisions.

At each cell, the routing information is constructed and maintained by communicating only with adjacent cells. Since the environment is initially unknown and can also change dynamically, it is essential to properly maintain this information throughout the lifetime of the system.

Centralized routing algorithms are obviously inadequate, so most known routing algorithms are inherently distributed, such as, distributed dynamic programming~\cite{bertsekas1982distributed,shoham2008multiagent}, or NetChange~\cite{tajibnapis1977correctness,tel2000introduction} to name just a few.

In this context, the notion of \emph{self-stabilization}~\cite{dijkstra1974self,altisen2019introduction} is particularly attractive due to its inherent robustness. Self-stabilizing algorithms are designed in a way that, starting from any possible global state (valid or invalid), the system is always guaranteed to reach a global valid state (or a cycle of valid states) after a finite number of transitions and then permanently remain in valid states in the absence of external influence (e.g., failures, state corruptions, topology changes).

In self-stabilizing routing~\cite{tel2000introduction,johnson2015safe}, a valid global state is one in which the combination of all routing tables defines a path from any cell to any other cell. Such algorithms are attractive because of their inherent adaptivity and robustness. Since AFADA is fully decentralized and entails many sources of uncertainty (e.g., cell addition/removal, message loss, crashes), its routing algorithm (Algo.~\ref{algo:routing}) is designed to be self-stabilizing.
The algorithm is inspired from classical self-stabilizing routing~\cite{tel2000introduction,johnson2015safe} but additionally copes with the addition/removal of cells at runtime.

\begin{algorithm}
  \caption{Update routing table (for cell $i$)}
  \label{algo:routing}
  \begin{algorithmic}
  \item $\mathit{dist}_i$: the estimated distance table
  \item $\mathit{next}_i$: the next-cell table
  \item $\mathcal{D}$: maximum diameter of $G$, constant value
  \end{algorithmic}
  \begin{algorithmic}[1]
    \State (\textit{repeat followings periodically})
    \State broadcast $\mathit{dist}_i$ to $i$'s neighbors
    \label{algo:routing:broadcast}
    \State clear $\mathit{dist}_i$, $\mathit{next}_i$, then $\mathit{dist}_i[i] \leftarrow 0$, $\mathit{next}_i[i] \leftarrow i$
    \label{algo:routing:clear}
    \For{each neighbor $j$ and each key $k$ in $\mathit{dist}_j$}
    \State \textbf{if}~$\mathit{dist}_j[k] \geq \mathcal{D}$~\textbf{then}~\textbf{continue}
    \If{$k \not\in \mathit{dist}_i.\mathit{keys}$ \textbf{or}
      $\mathit{dist}_j[k] + 1 < \mathit{dist}_i[k]$}
    \State $\mathit{dist}_i[k] \leftarrow \mathit{dist}_j[k] + 1$, $\mathit{next}_i[k] \leftarrow j$
    \EndIf
    \EndFor
    \label{algo:routing:endfor}
  \end{algorithmic}
\end{algorithm}

In Algo.~\ref{algo:routing}, each cell~$i$ maintains two tables:
$\mathit{dist}_i$ estimates the distance to other cells, and $\mathit{next}_i$ determines the next neighbor cell on the path to other cells.
Cell~$i$ periodically broadcasts $\mathit{dist}_i$ to its neighbors~\algoline{algo:routing:broadcast} (with \SI{2}{\second} period in our prototype), then updates both tables according to the minimal estimated distance of neighbor cells to target cells~\algolines{algo:routing:clear}{algo:routing:endfor}.

If $G$ remains unchanged for long enough, and cells $i$ and~$j$ are connected in $G$, then $\mathit{dist}_i[j]$ eventually holds the length of the shortest path connecting $i$ and~$j$.
Furthermore, for every cells $i,j,k$, if $i \neq j$ and $k = \mathit{next}_i[j]$ then $\mathit{dist}_k[j] = \mathit{dist}_i[j] - 1$. This implies that the routing tables correctly guide the robots.
The formal model and the proofs are beyond this paper.

\subsection{Integrated Planning and Execution}
\label{subsec:planning-execution}
Planning must ensure that robots' movements are collision-free, for which routing tables alone are insufficient.
To this end, it is important to also consider how the robots interact with the cells.

We present the basic behavior of robots and cells and their interaction, based on the time-independent model~\cite{okumura2020time} that copes with multiple agents on a graph moving towards their destinations without any timing assumptions.
The model is event-based in the sense that any change in the environment (e.g., start/end of movement, cell addition/removal, cell or robot failure, variable change) defines a new configuration (or global state).
In a distributed environment such as AFADA, the notion of ``simultaneity'' is difficult to realize~\cite{sheehy2015there}.
Robots should not rely on \emph{timings}, rather should rely on \emph{events} such as receiving messages or arriving at a new cell, and this justifies the use of a time-independent model.

Typically, the system repeats the following steps.
Before a robot can move, the cell requests and reserves the next cell on the path to the destination on behalf of the robot. The robot moves upon receiving confirmation from the cell, which is released by the next cell upon arrival of the robot. The reservation prevents collisions with other robots.
%

More precisely, assume that a robot $r$ is on a cell $i$ and its destination is $g$.
The procedure is as follows;
\begin{enumerate}
  \renewcommand{\labelenumi}{\arabic{enumi}.}
\item $r$ requests instructions from $i$ on how to go to $g$.
  \label{item:typical:init}
\item After receiving the request, $i$ enquires a neighbor cell $j$ about its availability (e.g., occupied, unoccupied).
  \label{item:typical:request}
\item $j$ replies its availability to $i$, according to whether another robot occupies $j$.
  If available, $j$ becomes \emph{reserved} by $r$.
\item If $j$ was unoccupied, continue from Step~\ref{item:typical:move} where $i$ instructs $r$ to move to $j$.
\item
  Otherwise, go back to Step~\ref{item:typical:request} with another $j$ or with the same $j$ but after waiting for some arbitrary time.
\item \label{item:typical:move}
    When $r$ receives the instruction from $i$, it starts moving towards $j$.
\item When $r$ arrives at $j$, it notifies $j$ which sends a release message to $i$ which is thus \emph{released}.
\item Repeat from Step~\ref{item:typical:init} with $j$ as the new $i$.
\end{enumerate}
In Step~\ref{item:typical:request}, $j$ is typically selected based on the routing table $\mathit{next}_{i}[g]$.
Alternatives include selecting $j$ randomly, in an attempt to break cycle of requests and thus potential deadlocks.

Variants can be considered, such as the multi-step reservation seen in the demo section (\S\ref{sec:demo:multistep}).
A comparison with other protocols is outside the scope of this paper.

\begin{figure*}[t]
  \centering
  \includegraphics[width=1\hsize]{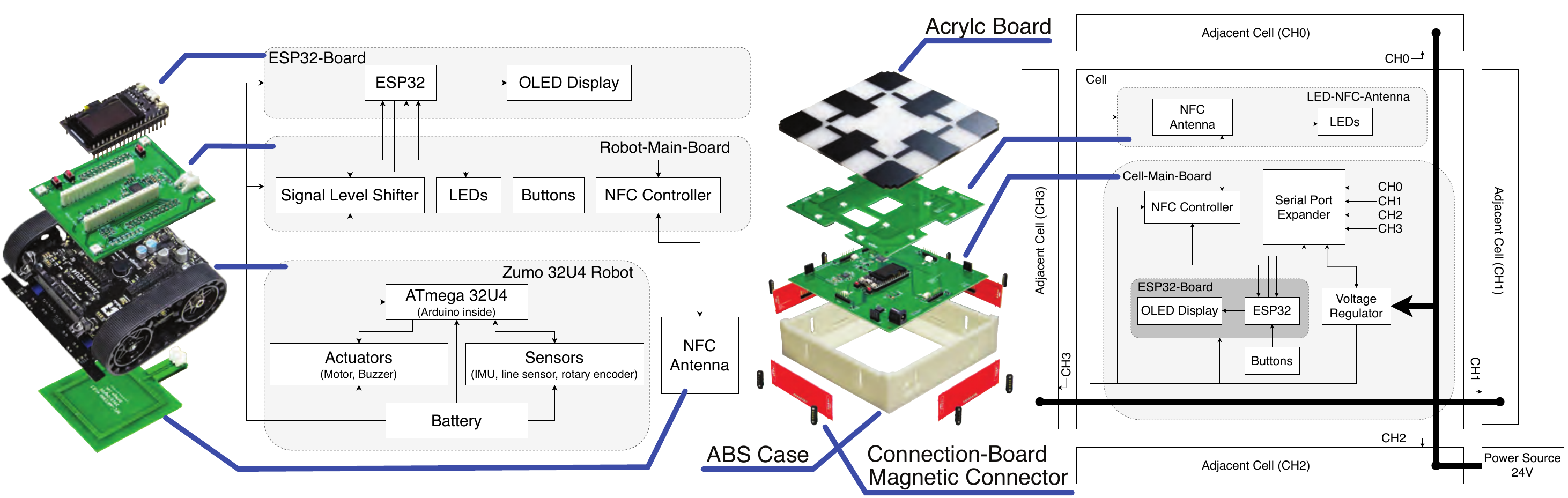}
  \caption{\textbf{Structures and block diagrams of a robot (left) and a cell (right).}}
  \label{fig:prototyping}
\end{figure*}

\section{Hardware Prototype}
\label{sec:prototyping}
This section describes the hardware implementation of robots (\S\ref{sec:prototyping:robot}) and cells (\S\ref{sec:prototyping:cell}).
The architecture is shown in Fig.~\ref{fig:prototyping}.
The robots use line tracing to move from a cell to the next.
The cells communicate with each other via a serial interface and with the robots via NFC (Near Field Communication) to avoid radio interference.
Cells can dynamically be attached or detached thanks to magnetic connectors.
Connections between cells supply power.

\subsection{Robot}
\label{sec:prototyping:robot}
\subsubsection{Requirements}
Robots need to identify an area of each cell to move from one cell to another cell correctly.
Robots need to communicate with underlying cells.

\subsubsection{Structure}
A robot consists of a Zumo 32U4 robot (Zumo) as a base with an interface board (Robot-Main-Board) holding an ESP32 microcontroller (ESP32-Board), an NFC controller (PN7150 chip), and LEDs.
An NFC antenna is placed under the robot.
The dimensions are \SI{100 x 100 x 42}~\SI{}{\cubic\mm}, and the weight is about \SI{290}{\gram}.

Functionally, the ESP32 works as the main controlling unit and runs FreeRTOS to support multi-tasking. The robot's motion is controlled by the ATmega microcontroller located on the Zumo.
Upon receiving instructions from the ESP32, the Zumo uses the line sensor array to follow patterns drawn on the cells and guide the robot to an adjacent cell.

\subsection{Cell}
\label{sec:prototyping:cell}
\subsubsection{Requirement}
A cell needs to communicate with robots on it.
It also communicates with adjacent cells, including a connection detection protocol to judge whether its neighbor exists or not.
It has sufficient space that stores one robot.
It is also critical to establish the power supply serving numerous cells while keeping the platform safe.

\subsubsection{Structure}
A cell consists of several printed circuit boards (Cell-Main-Board, LED-NFC-Antenna, Connection-Board), magnetic connectors, plastic parts (ABS Case), and an acrylic roof patterned for line tracing.
The mainboard (Cell-Main-Board) is equipped with the same ESP32 microcontroller as the robots, an NFC controller, and a serial port expander supporting four serial channels.
Magnetic connectors, which are installed on each side of the cell and connected to the mainboard via cables, are used for both communication between cells and shared power supply.
The cell has a size of \SI{160 x 160 x 41.5}~\SI{}{\cubic\mm} and weighs about \SI{440}{\gram}.
The average power consumption of a cell is about \SI{1.25}{\watt}.
We now consider the functional aspects of cells.

\paragraph{Cell-Cell Communication}
To send a message from one cell to another, first, the ESP32 microcontroller sends a message to a PIC (Peripheral Interface Controller; PIC24FJ64GA306-I/PT) in the cell, through SPI (Serial Peripheral Interface).
Then, the PIC relays the message to the adjacent cell via UART (Universal Asynchronous Receiver/Transmitter) communication.
Both UART and SPI are bidirectional serial interfaces.

\paragraph{Connection Detector}
Cells must detect the addition or removal of other cells on running to update the environmental information.
To monitor connection, the prototype uses two-stage detection: \emph{physical} and \emph{virtual}.
Physical detection uses an adapted version to UART from that of USB (Universal Serial Bus), detecting the connection by the voltage switching on the signal line as a trigger~\cite{specification2000universal}.
Virtual detection uses heartbeats~\cite{pasin2008failure}.
When two cells are actually connected, they exchange heartbeats periodically (period of \SI{3}{\second} in the demo).
When the cell fails to receive heartbeats for a certain time (timeout \SI{10}{\second} in the demo) from some channel, the cell regards this channel as disconnected.

\paragraph{Power}
There are two ways to power a cell: a)~connecting the cell directly to a power adapter (supply voltage: \SI{24}{\volt}), or, b)~connecting the cell to a powered cell.
The system can be powered from several power adapters (e.g., 6 adapters for 100 cells).
The \SI{24}{\volt} line is connected to an onboard DC/DC converter, which provides a \SI{5}{\volt} line used to supply power to the components in the cell.
For safety, the prototype includes a switching circuit with relays so that the \SI{24}{\volt} line, dangerous to humans, is not live unless an adjacent cell is connected.

{
  \newcommand{\colwidth}{0.3\hsize}
  \newcommand{\rowspace}{-0.2cm}
  \newcommand{\instance}[1]{{#1}\vspace{0.05cm}\\}
  \newcommand{\photowidth}{0.88\hsize}
  \begin{figure*}[ht!]
    \centering
    \begin{tabular}{ccc}
      \begin{minipage}{\colwidth}
        \centering
        \begin{tabular}{c}
          \begin{minipage}{\photowidth}
            \centering
            \instance{\insone}
            \includegraphics[width=1\hsize]{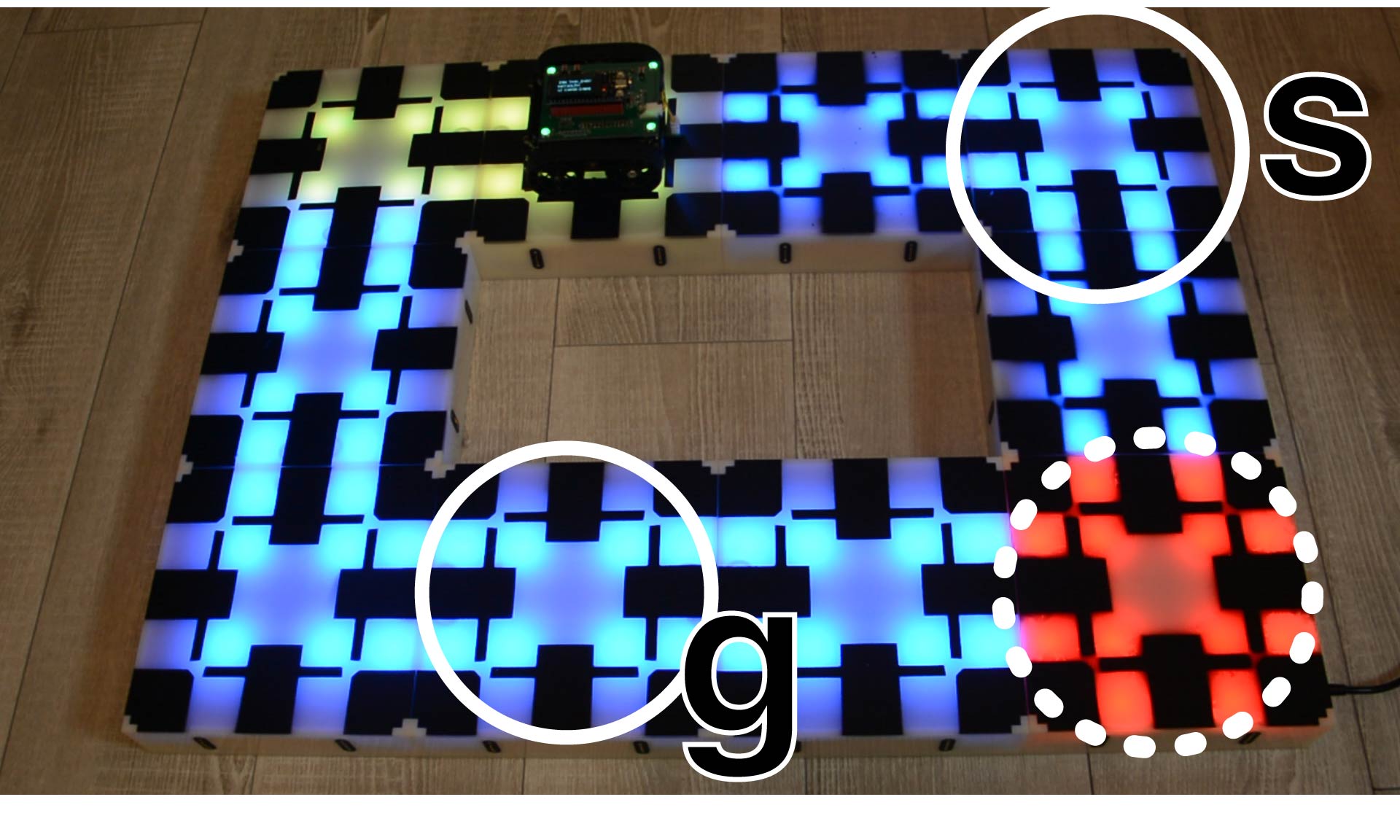}
            \vspace{\rowspace}
          \end{minipage}
          \\
          \begin{minipage}{1\hsize}
            \centering
            \includegraphics[width=1\hsize]{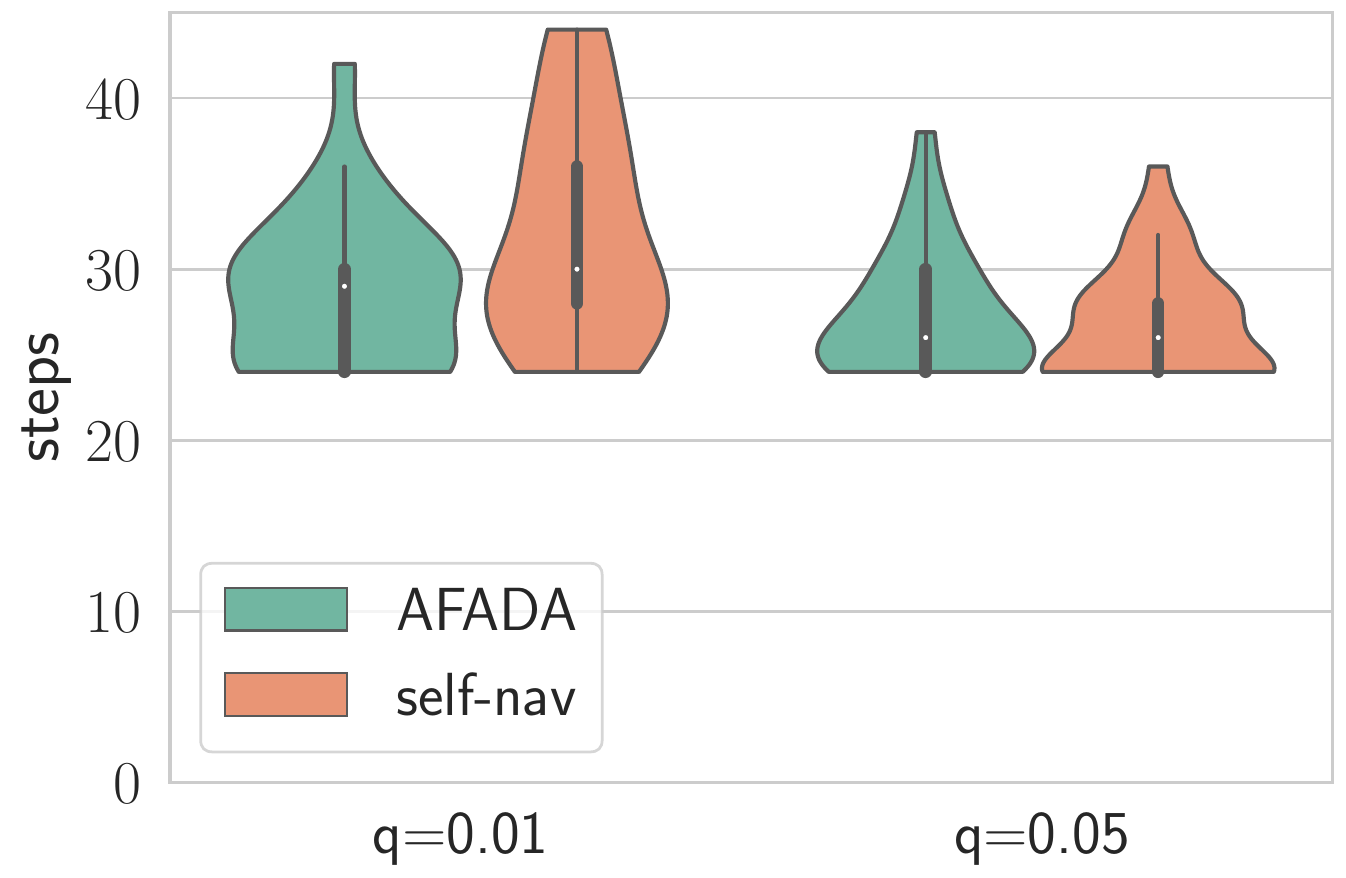}
          \end{minipage}
        \end{tabular}
      \end{minipage}
      &
      \begin{minipage}{\colwidth}
        \centering
        \begin{tabular}{c}
          \begin{minipage}{\photowidth}
            \centering
            \instance{\instwo}
            \includegraphics[width=1\hsize]{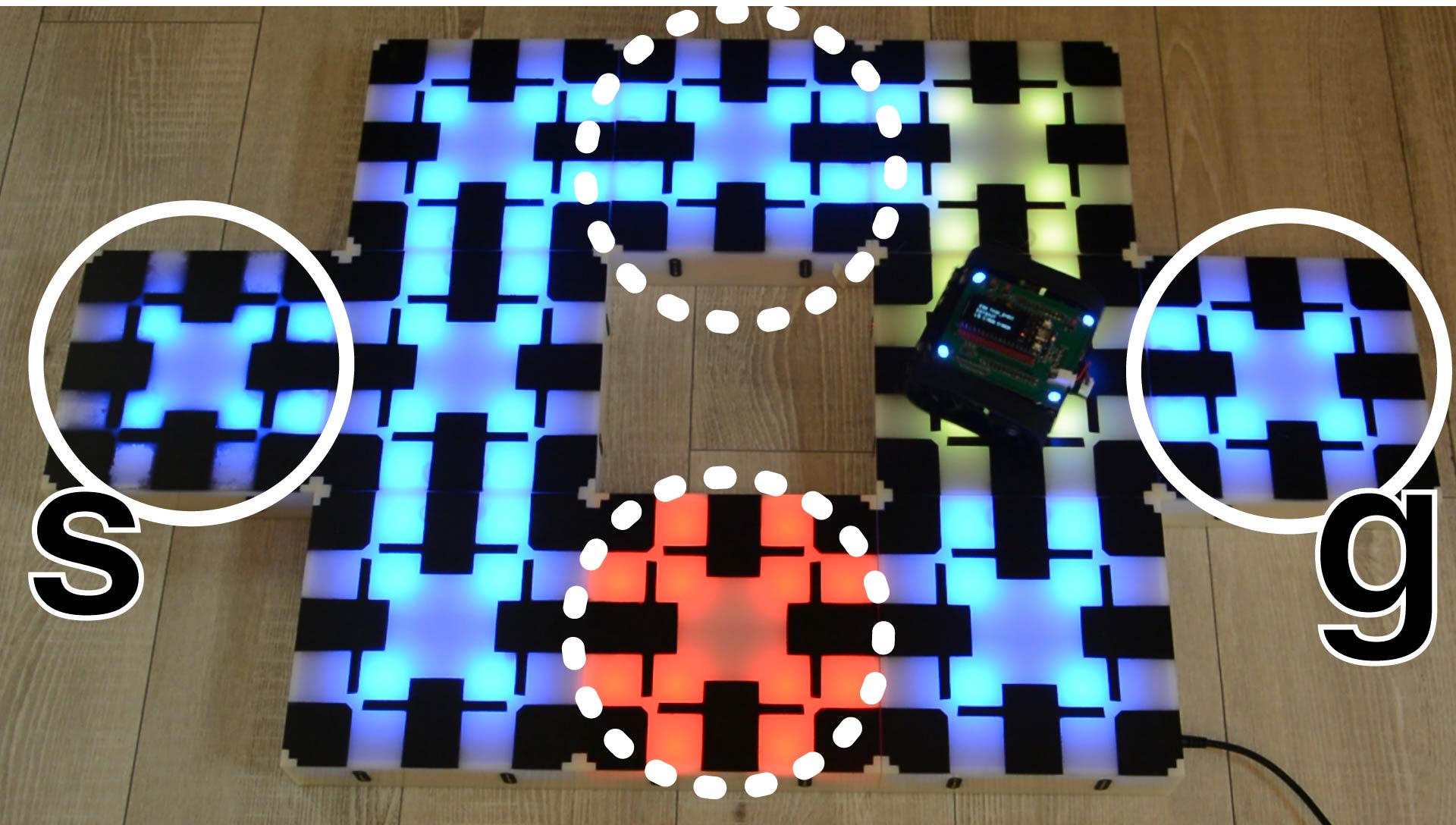}
            \vspace{\rowspace}
          \end{minipage}
          \\
          \begin{minipage}{1\hsize}
            \centering
            \includegraphics[width=1\hsize]{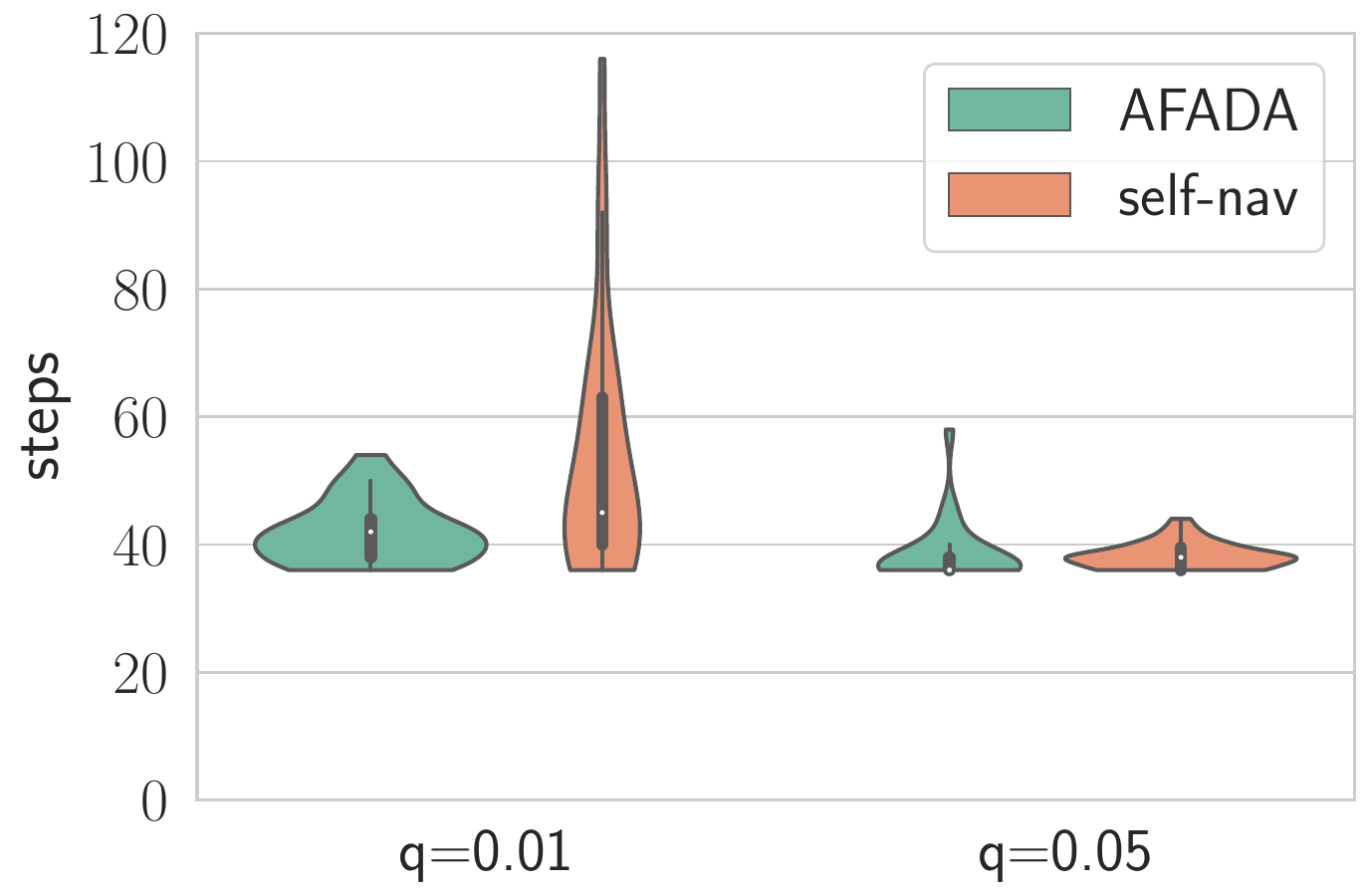}
          \end{minipage}
        \end{tabular}
      \end{minipage}
      &
      \begin{minipage}{\colwidth}
        \centering
        \begin{tabular}{c}
          \begin{minipage}{\photowidth}
            \centering
            \instance{\insthree}
            \includegraphics[width=1\hsize]{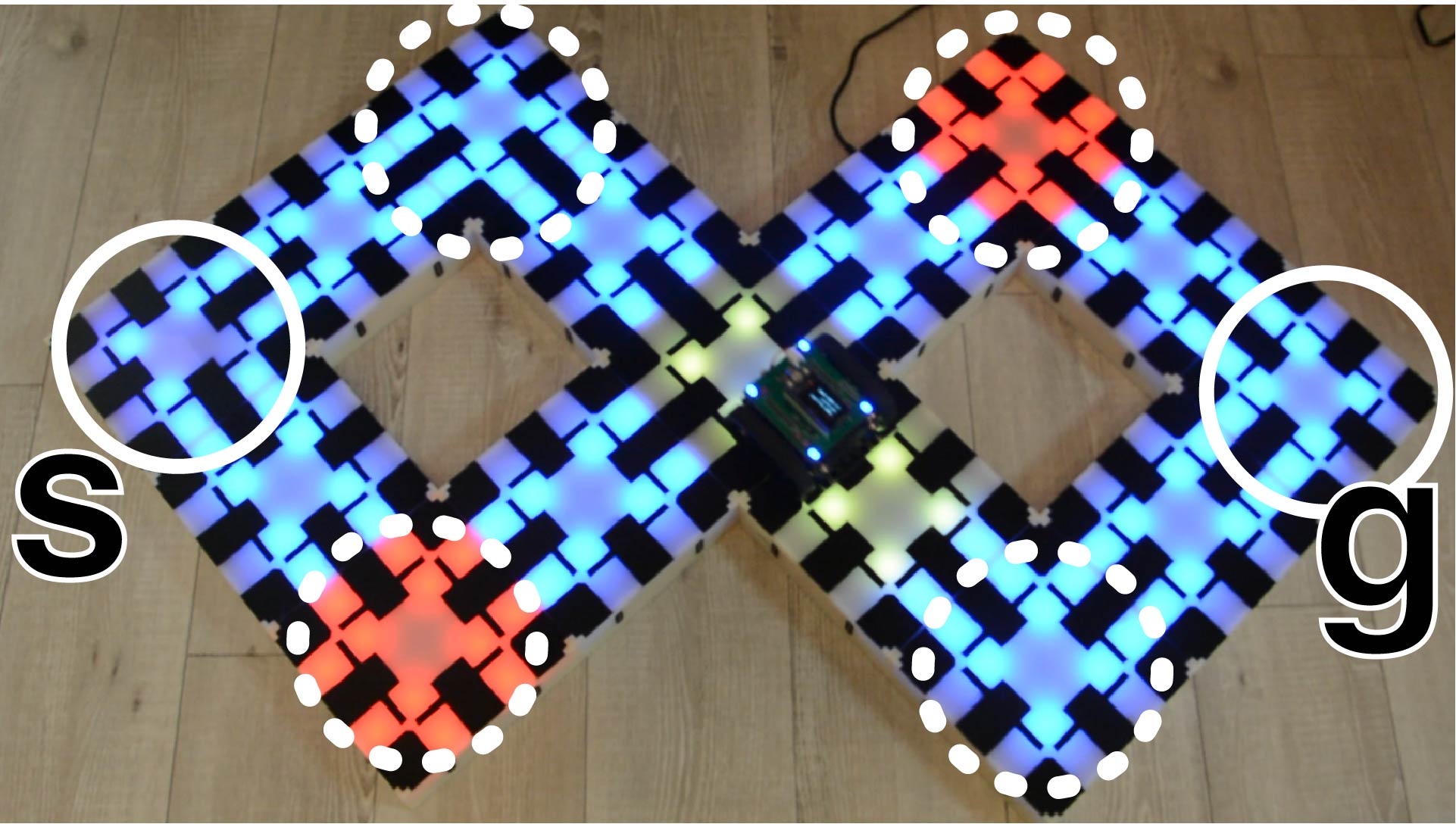}
            \vspace{\rowspace}
          \end{minipage}
          \\
          \begin{minipage}{1\hsize}
            \centering
            \includegraphics[width=1\hsize]{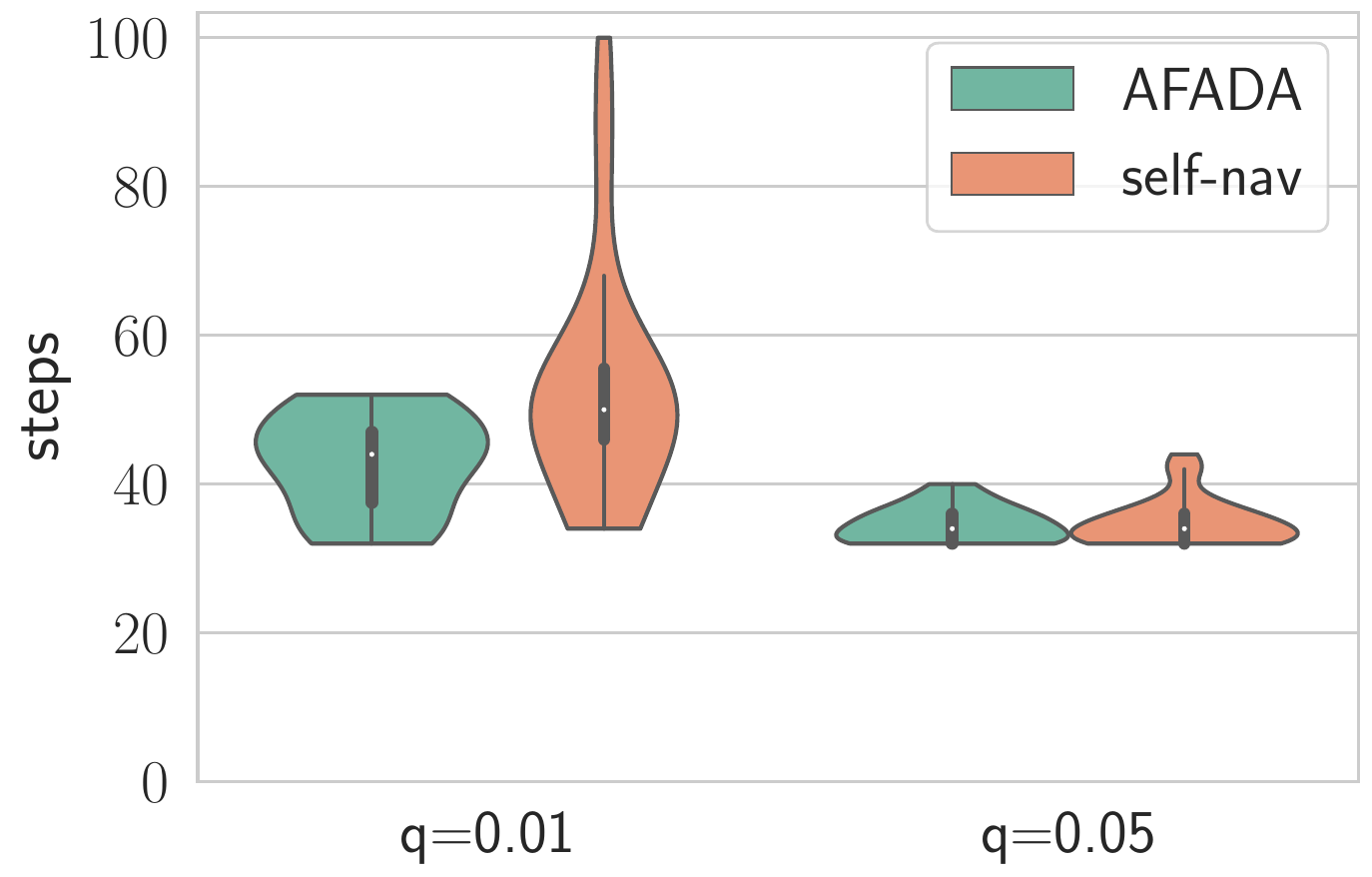}
          \end{minipage}
        \end{tabular}
      \end{minipage}
  \end{tabular}
  \caption{\textbf{The results of navigation tasks in a dynamic environment.}
    Target cells (``s'' and ``g'') are identified by a solid circle, and potentially failed cells are identified by a dashed circle and highlighted in red when their status is actually failed.
  }
  \label{fig:result-nav}
\end{figure*}
}

\section{Evaluation}
\label{sec:evaluation}
To evaluate the idea of offloading representation and planning, we first consider a single robot navigating in a dynamic environment.
As baseline, we used a robot navigating by itself while collecting information about the environment.

\subsection{Task}
The task consists of multiple round-trips between two cells.
To emulate a dynamic environment, several cells stochastically switch their status between \emph{correct} and \emph{failed}.
Failed cells cease to communicate and robots cannot pass through them.
Surrounding cells of the failed cell recognize a disconnection to the cell, resulting in an update of the routing tables.
Every second, correct cells change their status to failed with a probability $p$ if no robot is present.
Failed cells change their status to correct with a recovery probability $q$.
The metric counts the total number of steps taken by the robot (a step is a movement from a cell to a neighbor). Fewer steps is better and means less overhead.

Three carefully designed fields (\insone, \instwo, and \insthree) were used, shown in Fig.~\ref{fig:result-nav}, which includes annotations of target cells and potentially failed cells.
The numbers of round trips are three times both in \insone and in \instwo, and two times in \insthree.
Even though one cell crashes, all fields have detours for round trips, hence moves approaching failed cells result in unnecessary steps.
In \instwo and \insthree, target cells are potentially disconnected by two failed cells;
the robot should not move during disconnection to avoid unnecessary steps.
We fixed $p=0.01$ but set $q$ in two conditions, $0.01$ or $0.05$, then run the robot with 30 repetitions for each condition.

\subsection{Comparison}
As a baseline for comparison, we tested a robot navigating itself according to its internal map, using the Zumo robot base.
Assume that the robot does not know the availability of locations until it approaches.
In this experiment, the robot can detect failed cells only when adjacent to them.
The status of cells is simulated in the robot.

The robot first plans the path for the trip (outward or return) using the shortest path algorithm, then moves.
When it meets a failed cell on its way, it replans the path around it, and resumes movement.
Herein, we call this style as \selfnav.
Note that \selfnav never interacts with active cells.

\subsection{Results}
The results are summarized in Fig~\ref{fig:result-nav}.
In general, AFADA achieved efficient robot moves compared to \selfnav when the recovery probability $q$ was small, i.e., failed cells existed frequently.
We describe the analysis as follows.
The statistical significance threshold was set to~$0.05$.

\subsubsection{\insone}
AFADA failed to complete the task four times when $q=0.05$ and three times when $q=0.01$ due to message loss of the reservation protocol between cells.
When $q=0.05$, median scores were both $26$ in AFADA and \selfnav;
the distributions in the two groups did not differ significantly (Mann–Whitney $U=362.5$, $n_1=26$, $n_2=30$, $P=0.24$, two-tailed).
When $q=0.01$, median scores were $29$ in AFADA and $30$ in \selfnav;
we observed significant difference ($U=283.0$, $n_1=27$, $n_2=30$, $P=0.03$).

\subsubsection{\instwo}
AFADA failed several times.
When $q=0.05$, median scores were $36$ in AFADA and $38$ in \selfnav;
the distributions did not differ significantly ($U=298.0$, $n_1=25$, $n_2=30$, $P=0.08$).
When $q=0.01$, median scores were $42$ in AFADA and $45$ in \selfnav;
there was a significant difference ($U=258.5$, $n_1=27$, $n_2=30$, $P<0.01$).

\subsubsection{\insthree}
In this case, AFADA failed in almost half of the trials; 12 times when $q=0.05$ and 14 times when $q=0.01$.
When $q=0.05$, median scores were both $34$ in AFADA and in \selfnav;
the distributions did not differ significantly ($U=269.5$, $n_1=18$, $n_2=30$, $P=0.5$).
When $q=0.01$, median scores were $44$ in AFADA and $50$ in \selfnav;
we observed significant difference ($U=135.0$, $n_1=16$, $n_2=30$, $P<0.01$).

\subsection{Discussion}
AFADA can achieve efficient robot moves because the environment ``knows'' which cells are available before the robot moves.
Note that the environmental update also requires time;
we observed that AFADA did not always result in optimal moves.

AFADA failed sometimes due to message loss.
Our prototype allows operations of cell addition or removal physically, however, as a disadvantage, the connection might be faulty.
As seen in \insthree, when the number of cells increases, this problem cannot be neglectable.
Solutions include developing robust message channels or self-stabilizing planning and execution protocols, which we plan for future work.

\section{Case Studies}
\label{sec:case-studies}
This section presents demos showing the potential of AFADA.
A supplementary movie%
\footnote{
  Available on \url{https://dfg-lab.github.io/afada/}
}
presents them as well as other demos.

\subsection{Navigation in Reconfigurable Environment}
\begin{figure}[t]
  \centering
  \includegraphics[width=1\hsize]{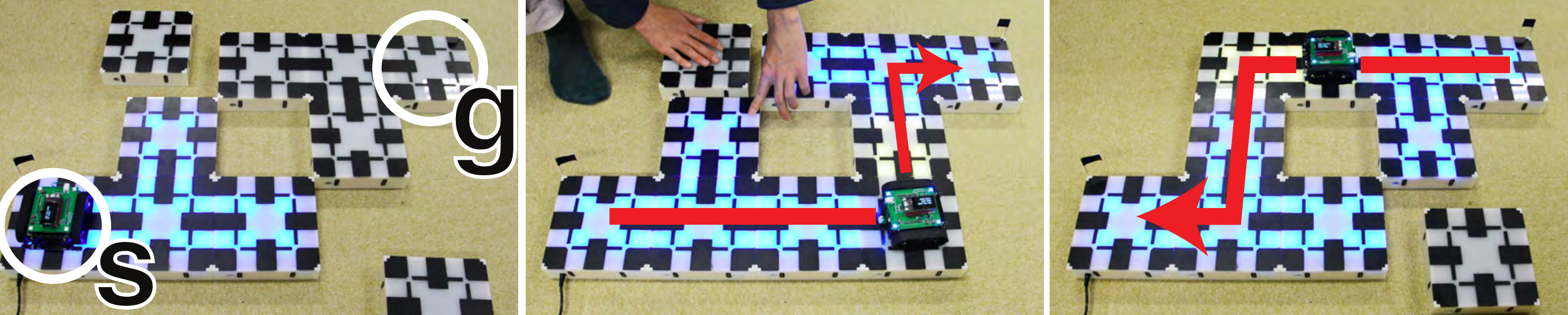}
  \caption{\textbf{Demo of navigation in a reconfigurable environment.}
    Two cells are disconnected initially \emph{(left)}. The robot moves once the bottom one is connected \emph{(center)}, and then returns via a different route \emph{(right)}.
  }
  \label{fig:demo-nav-dynamic}
\end{figure}

As illustrated in Fig.~\ref{fig:demo-nav-dynamic}, cells can guide robots correctly in spite of environmental changes due to adding/removing cells.
A task is a round trip between two cells $s$ and $g$.
Cells are added or removed during the execution.
At first, $g$ is unreachable and the cell informs the robot to wait in place.
The robot starts moving after the cells detect the connection of $g$.
Next, we added a new cell to create a new path, while removing a cell previously used by the robot.
The environment immediately updates the routing tables, and successfully guides the robot back to $s$ via the new path.
This illustrates the self-reconfiguration of the system in the face of physical changes in the environment.

\subsection{Multi-robot Navigation}
In multi-robot scenarios, colliding is fatal;
AFADA provides collision-free moves of robots.

\subsubsection{Multi-robot Path Finding and Execution}
\label{sec:demo:multistep}
Initially, all robots have distinct locations and destinations.
Fig.~\ref{fig:demo-nav-multi} shows a configuration used in the demo.
A solution is to make robots move to their destinations.
In addition to the single-step reservation protocol in Section~\ref{subsec:planning-execution}, this demo uses a multi-step reservation;
let cells reserve multiple cells ahead before a robot moves.
When vacant cells receive a request from neighbors, they \emph{forward} the request to the next adjacent cells.
This forwarding of the request continues until the request is rejected by an occupied cell.
Upon receiving a rejection, these cells then relay acknowledgments in the reverse order.
As a result, a robot reserves several cells ahead and releases them one-by-one as it moves.
Both single-step and multi-step reservation styles used a random choice of cells when receiving a rejection of the request to break deadlocks.

\subsubsection{Automated Parking}
As an application, we emulate an automated parking system (Fig.~\ref{fig:demo-nav-multi}).
The parking lot is modeled as a $4\stimes 4$ grid.
Cells in the leftmost/rightmost columns are parking space, and the two center columns are aisle.
Robots can enter or leave the parking lot from the bottom row of the environment.
These cells and the aisle are programmed to guide robots in one-way traffic using the reservation protocol to avoid collisions.
The robot greedily looks for a vacant parking space following the one-way traffic.
If the robot successfully enters a vacant parking space, it waits for a while
(corresponding to the driver running an errand).
Then, the robot leaves the parking lot, heads to the exit, and leaves the environment.

{
  \newcommand{\colwidth}{0.45\hsize}
  \begin{figure}[t]
    \centering
    \begin{tabular}{cc}
      \begin{minipage}{\colwidth}
        \centering
        \includegraphics[width=1\hsize]{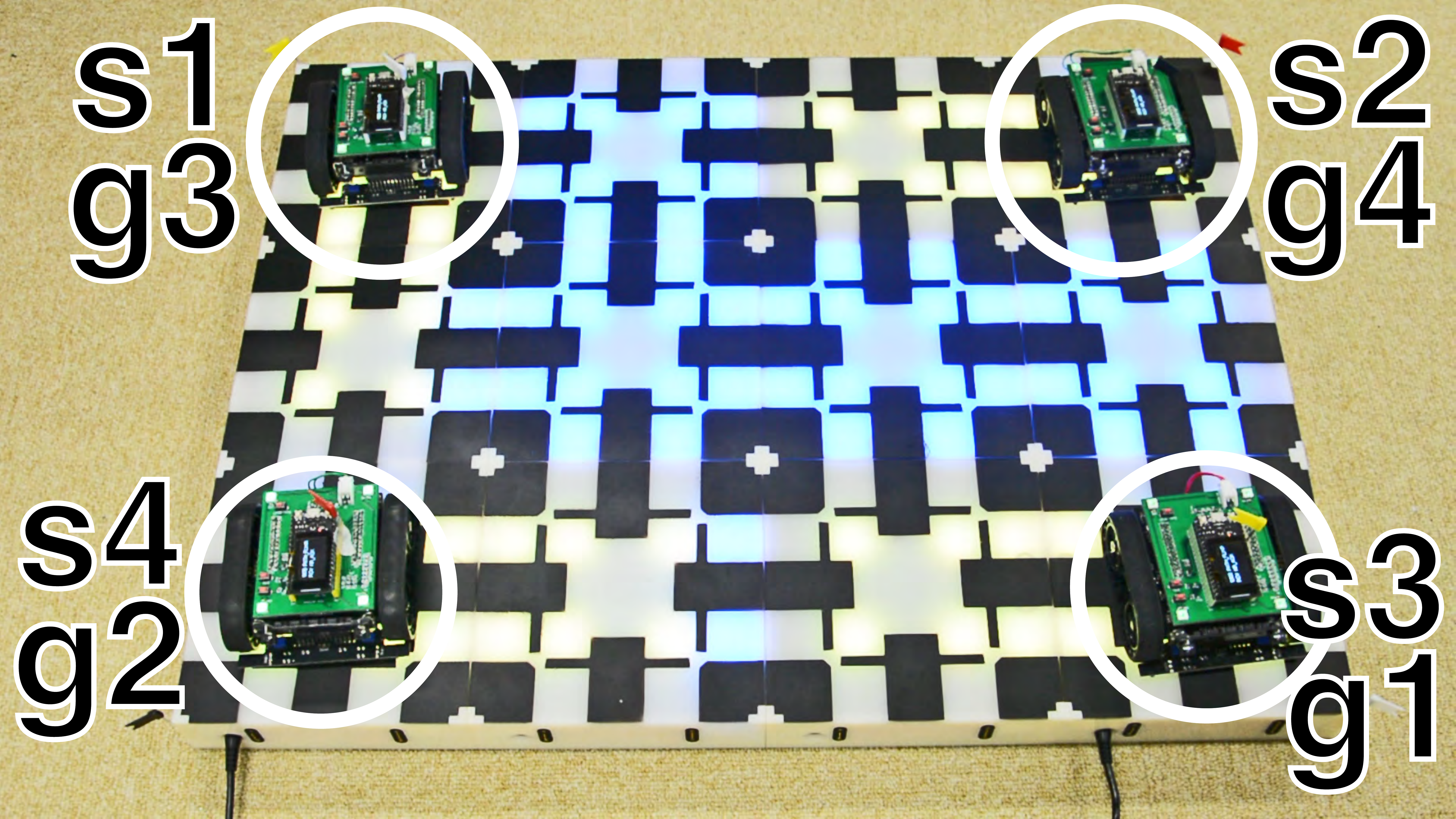}
      \end{minipage}
      &
      \begin{minipage}{\colwidth}
        \centering
        \includegraphics[width=1\hsize]{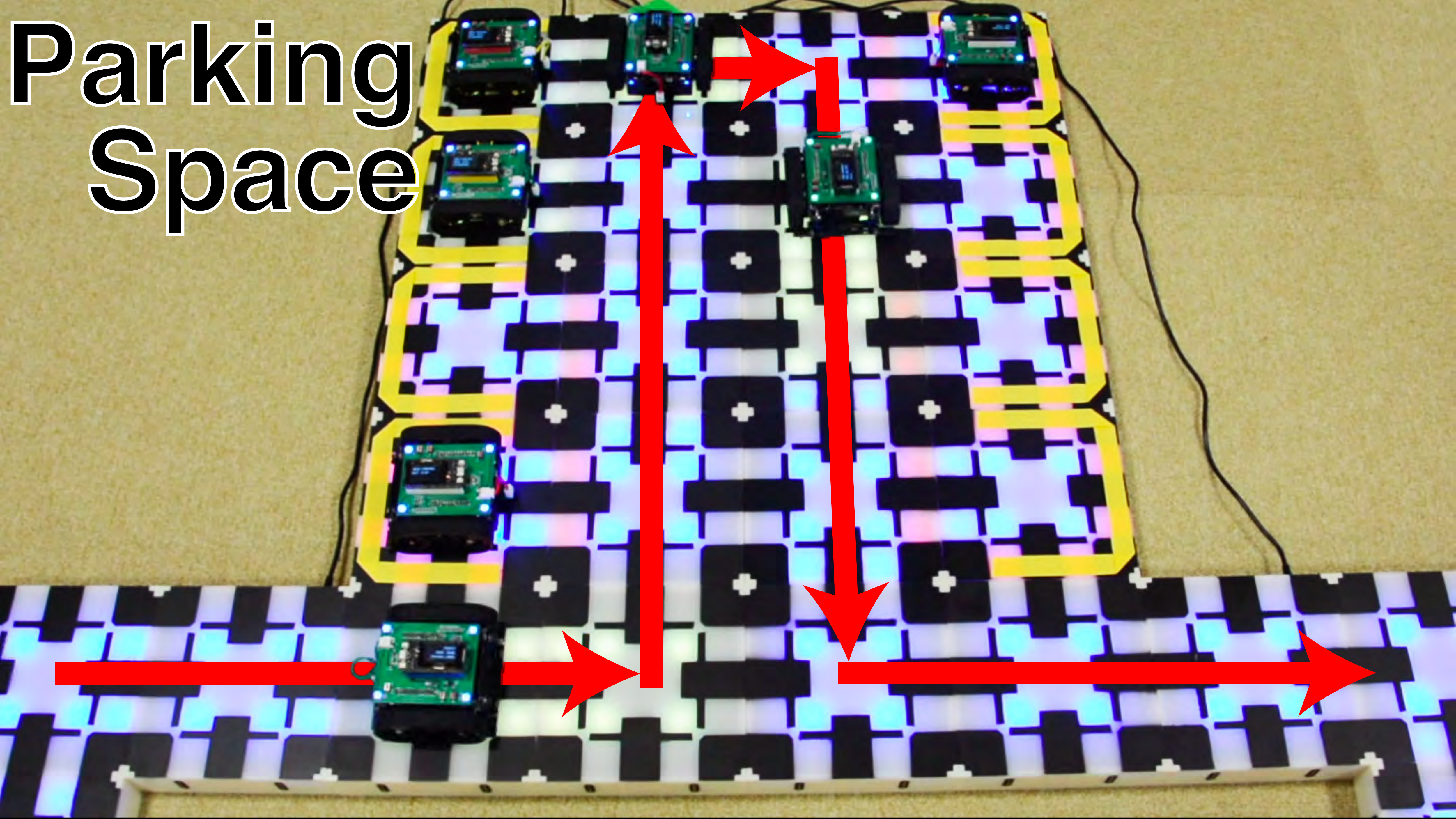}
      \end{minipage}
    \end{tabular}
    \caption{\textbf{Demo of navigation for multiple robots.}
      \emph{(left)} the pathfinding and execution task are shown, with start (``s'') and goal (``g'') locations.
      \emph{(right)} automated parking is shown with red arrows corresponding to the one-way aisle. The cells with yellow tapes are parking lot.
    }
    \label{fig:demo-nav-multi}
  \end{figure}
}

\section{Discussion and Conclusion}
\label{sec:discussion}
We presented a novel fully decentralized scheme with robot-environment interactions for robot navigation, with AFADA and its implementation as proof-of-concept.
The key concept consists in offloading both representation and planning from robots to the environment with cells as computing units embedded spatially.
Robots do not plan their moves alone, rather, cells communicate with neighboring cells and collectively take charge of robots' movements.
In the experiment, we demonstrated that AFADA achieved efficient robot moves using the navigation task in the dynamic environment, followed by several demos such as pathfinding and execution for multiple robots.
We expect that the concept can be applied to many domains of multi-robot systems as infrastructure, such as automated warehouses, smart parking, and the cooperative behavior of autonomous cars.

Quite a few challenges remain to apply the concept to realistic scenarios, e.g., introducing an active environment on a large scale is costly.
One important issue is that the system is inherently distributed and fully decentralized.
This implies that we cannot expect a perfect execution devoid of any unexpected events such as the failure of or an increase in system components.
An example can be seen in navigation experiments of \insthree in Section~\ref{sec:evaluation}.
This is where paradigms and principles of distributed computing are attractive, just like we used self-stabilizing algorithms.

Future work includes the following.
1)~\emph{Develop cell-driven path planning algorithms}.
We only presented a basic protocol to prevent collisions between robots, however, it does not yet ensure desirable properties such that all robots always reach their destinations within a finite time.
Furthermore, there is considerable room for improvement in the efficiency of trajectories for multiple robots.
2)~\emph{Address large robots}.
We currently limit the size of robots to be smaller than a cell for simplicity.
Addressing large robots that occupy several cells simultaneously, is fruitful for coordination between heterogeneous robots.
This is closely related to diagonal or any-angle moves of robots in a grid environment.


\section*{Acknowledgments}
This work was partly supported by the NTT FACILITIES Collaborative Research Project.
The authors thank Mattias Evaldsson, Daniel Gst\"{o}hl, and Shoma Mori for their support in the early phases of development; as well as Kazuya Iimuro, Haruka Katahira, Wataru Kinota, and Xin Cen, for their help with building the numerous cells.


\bibliographystyle{IEEEtran}
\bibliography{ref}

\end{document}